\documentclass[10pt,twocolumn,letterpaper]{article}

\usepackage[pagenumbers]{cvpr} 

\usepackage[dvipsnames]{xcolor}
\usepackage{amsmath}
\usepackage{verbatim}
\usepackage{tabularray}
\usepackage{rotating}
\usepackage{array}
\usepackage{multirow}
\usepackage{ragged2e}
\usepackage{scalerel}
\usepackage{graphicx}
\usepackage{wrapfig}
\usepackage{booktabs}
\usepackage{cite}
\usepackage{amsmath}
\usepackage{algorithm}
\usepackage{xspace}
\usepackage{algpseudocode}
\usepackage{mathtools}
\usepackage{breqn}
\usepackage[accsupp]{axessibility}
\usepackage{adjustbox}

\DeclarePairedDelimiter\floor{\lfloor}{\rfloor}

\newcommand{\ubar}[2]{\mathbf{{#1}}_{#2}}
\newcommand{\red}[1]{{\color{red}#1}}

\newcommand{\subheader}[1]{\vspace{0.25\baselineskip} \noindent \textbf{#1.}}
\newcommand{\citethis}{\colorbox{yellow}{\makebox(17, 2.5){\color{red}\textbf{[cite]}}}}

\newcommand{\Reasoner}{Expert Knowledge Fusion Network\xspace}

\newcommand{\ee}{expert embedders\xspace}
\newcommand{\Ee}{Expert Embedders\xspace}

\newcommand{\Algname}{Ensemble of Expert Embedders\xspace}

\newcommand{\gensetbl}{\mathcal{G}_0}
\newcommand{\genset}{\mathcal{G}_k}
\newcommand{\prevgenset}{\mathcal{G}_{k-1}}

\newcommand{\membuf}{\mathcal{M}_k}
\newcommand{\prevbuf}{\mathcal{M}_{k-1}}

\newcommand{\newdata}{\mathcal{D}_k}

\newcommand{\synthset}{\mathcal{S}_k}
\newcommand{\prevsynthset}{\mathcal{S}_{k-1}}

\newcommand{\realset}{\mathcal{R}}

\newcommand{\expset}{\Phi_k}

\newcommand{\embseq}{\{\mathbf{x}_0,\ldots,\mathbf{x}_k\}}
\newcommand{\embseql}{\{\mathbf{x}^\ell_0,\ldots,\mathbf{x}^\ell_k\}}
\newcommand{\embseqlarg}{\mathbf{x}^\ell_0,\ldots,\mathbf{x}^\ell_k}

\newcommand{\pullupp}{\vspace{-0.4\baselineskip}}
\newcommand{\pulluppp}{\vspace{-0.8\baselineskip}}

\newcommand{\skipless}{
	\setlength{\abovedisplayskip}{3.75pt}
	\setlength{\belowdisplayskip}{3.75pt}
	\setlength{\abovedisplayshortskip}{3.75pt}
	\setlength{\belowdisplayshortskip}{3.75pt}
}
\newcommand{\skipnormal}{
	\setlength{\abovedisplayskip}{8pt}
	\setlength{\belowdisplayskip}{8pt}
	\setlength{\abovedisplayshortskip}{5pt}
	\setlength{\belowdisplayshortskip}{8pt}
}

\newcommand{\pluseq}{\mathrel{+}=}

\definecolor{cvprblue}{rgb}{0.21,0.49,0.74}
\usepackage[pagebackref,breaklinks,colorlinks,citecolor=cvprblue]{hyperref}

\def\paperID{0007} 
\def\confName{CVPR}
\def\confYear{2024}

\title{E3: Ensemble of Expert Embedders for Adapting Synthetic Image Detectors to New Generators Using Limited Data}

\author{\vspace{0.1cm} Aref Azizpour\quad Tai D. Nguyen\quad Manil Shrestha\quad Kaidi Xu\quad Edward Kim\quad Matthew C. Stamm\\
{Drexel University}\\
Philadelphia, PA, USA\\
{\tt\small \{aa4639, tdn47, ms5267, kx46, ek826, mcs382\}@drexel.edu}\\
{\footnotesize \vspace{0.1cm} Code: \url{https://github.com/ArefAz/E3-Ensemble-of-Expert-Embedders-CVPRWMF24}}
}

\begin{document}
\maketitle
\begin{abstract}

As generative AI progresses rapidly, new synthetic image generators continue to emerge at a swift pace.
Traditional detection methods face two main challenges in adapting to these generators: 
the forensic traces of synthetic images from new techniques can vastly differ from those learned during training, 
and access to data for these new generators is often limited.
To address these issues, we introduce the Ensemble of Expert Embedders (E3), a novel continual learning framework for updating synthetic image detectors. E3 enables the accurate detection of images from newly emerged generators using minimal training data.
Our approach does this by first employing transfer learning to develop a suite of expert embedders, each specializing in the forensic traces of a specific generator. 
Then, all embeddings are jointly analyzed by an Expert Knowledge Fusion Network to produce accurate and reliable detection decisions.
Our experiments demonstrate that E3 outperforms existing continual learning methods, including those developed specifically for synthetic image detection. 
\end{abstract}


\section{Introduction}



Over recent years, a number of AI-based techniques have been developed to create visually realistic synthetic images.  While these synthetic image generators can be used for creative or artistic purposes, they can also be used to malicious ones.  Specifically, they enable the creation of fake images that can be used for misinformation or disinformation.

\begin{figure}
    \centering
    \includegraphics[width=0.9\linewidth]{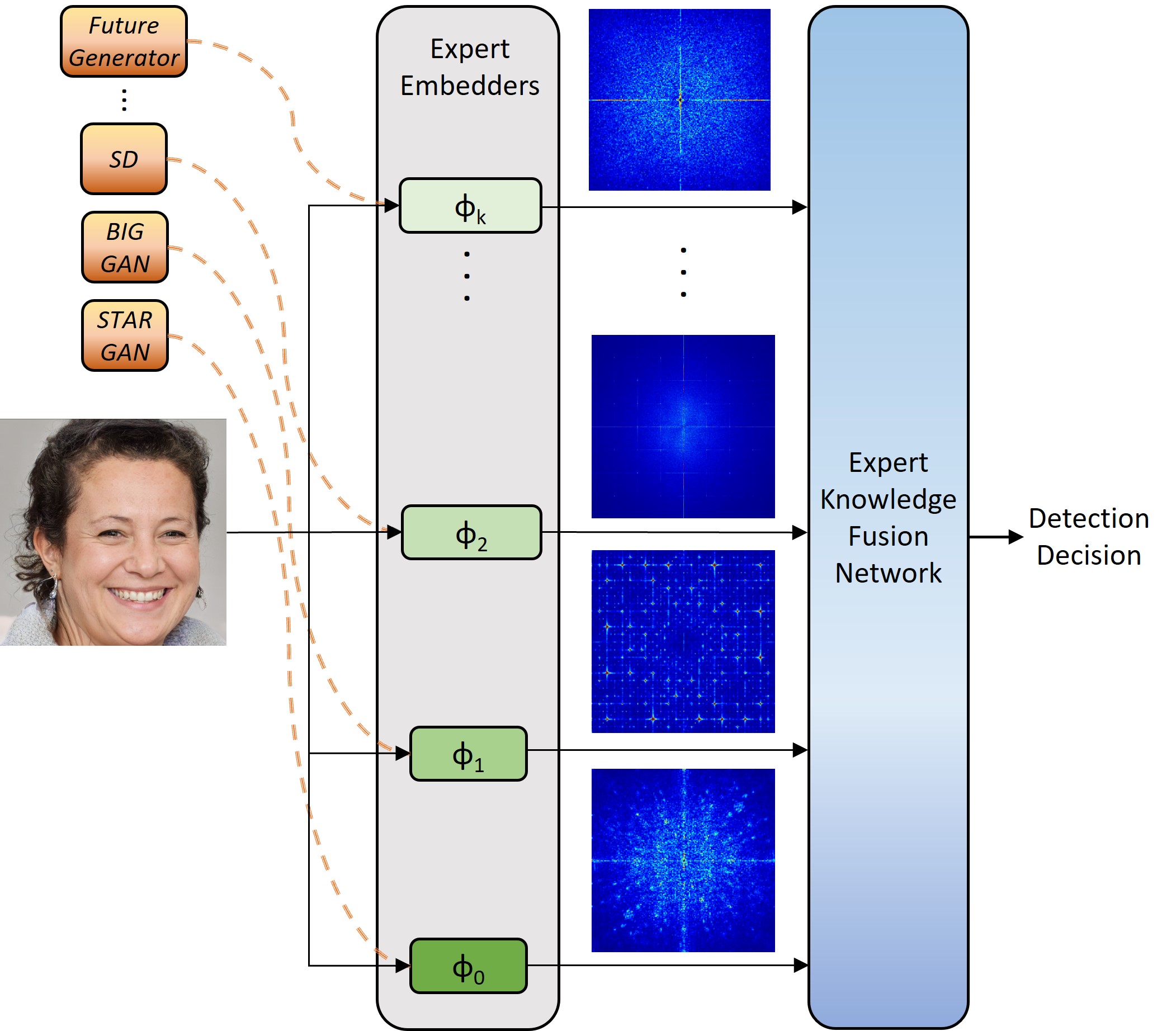}
    \caption{Visualization of the Ensemble of Expert Embedders (E3) framework aimed at enhancing synthetic image detection in response to new generators. Examples of the forensic residual traces for different generator architectures are also depicted.}
    \label{fig:front_page_graphics}

    \pulluppp\pullupp
\end{figure}
To address these challenges, considerable research efforts have been directed towards the development of synthetic image detection techniques. Prior work has demonstrated the effectiveness of forensic neural networks in discerning synthetic images by identifying unique traces left behind by different image generators. However, a significant limitation of existing detectors arises when they encounter images generated by previously unseen or emerging techniques, whose traces differ substantially from those in the training data.

This poses a critical need for continually updating synthetic image detectors to adapt to new generators. However, this task presents several challenges. Traditional approaches to updating detectors often encounter issues such as catastrophic forgetting and the impracticality of storing and retraining on large datasets~\cite{LWF, ER, udil}. Moreover, limited data availability from newly emerging generators, especially those not yet publicly accessible, further complicates the updating process.

In this paper, we propose a novel approach to address these challenges. Instead of relying on a single network to capture all forensic traces, we introduce an Ensemble of Expert Embedders (E3) framework. Each expert embedder is created through transfer learning and specializes in capturing traces from a specific image generator. These expert embedders collectively generate a sequence of embeddings, which are then analyzed by an Expert Knowledge Fusion Network (EKFN). The EKFN leverages information from all expert embedders to accurately perform synthetic image detection.
Our experimental results demonstrate that the proposed E3 framework significantly outperforms existing continual learning approaches, including those designed specifically for updating synthetic image detectors, 
and can perform strongly even with very limited data from new generators. 
%
%
The novel contributions of this paper are:
\begin{itemize}
    \item We propose E3, a new approach that can update synthetic image detectors  to accurately detect newly emerging generators, while requiring only minimal amounts of training data to be retained in a memory buffer. 

    \item We develop a novel approach to create a set of expert embedders to accurately capture traces from each new target generator.  Each expert can be adapted using a small set of images from its target generator.

    \item We propose an expert knowledge fusion network able to examine forensic evidence produced by the set of all experts and make accurate detection decision.

    \item We conduct an extensive set of experiments demonstrating that E3 outperforms competing approaches from continual learning, including approaches specifically designed to update synthetic image detectors.   Notably, E3 consistently outperforms competitors across various detector architectures and excels even with limited data from new generators.

    

\end{itemize}

\section{Background}

\subheader{Synthetic Image Generators}
Considerable effort has been devoted to developing systems that can visually understand the world through the process of mimicking; the first of such work is Variational Auto-Encoder by Kingma and Welling~\cite{kingma2013auto}, which led to the development of Generative Adverserial Networks (GANs) by Goodfellow~et al.~\cite{goodfellow2014generative}. This work inspired many other subsequent works~\cite{brock2018large, mirza2014conditional, zhu2017unpaired, isola2017image, karras2019style}, which continued to improve visual understanding through enhancing the generation's quality, diversity, and realism. Notably, the introduction of diffusion model for image generation by Ho~et al.~\cite{ho2020denoising} set the stage for the explosion of research in this area, resulting in many popular generation methods such as: Stable Diffusion~\cite{rombach2022high}, DALL·E~\cite{ramesh2021zero}, Midjourney~\cite{midjourney}, Cascade Diffusion~\cite{cascade_diff}, etc.~\cite{controlnet, diff_beat_gans, cond_text_im_gen_diff, im_gen_3d_diff}. Due to the rapidly increasing rate of emergence of new generation methods, existing synthetic image detectors face a formidable challenge. These detectors often have limited capabilities, restricting them to detecting only the generators seen during training~\cite{deepfake_and_beyond,corvi2023detection}.


\subheader{Synthetic Image Traces and Detection}
Numerous efforts have been undertaken to distinguish synthetic images from real ones. Marra~et al.~\cite{marra2019gans} demonstrated that GAN-generated images possess unique ``fingerprints'' useful for detection and source attribution. This work showed that individual image generators all left behind identifiable artifacts, called forensic traces, that can be used to detect their generated images. While prior work has examined transferable forensic features~\cite{mayer2018learning}, subsequent research
 demonstrated the difficulty for a detector to generalize to forensic traces from a new family of generators~\cite{corvi2023detection, may2023comprehensive}. To address the rapid release of generators, a broad spectrum of new synthetic image detectors, and more recently, synthetic video detectors, were developed~\cite{corvi2023detection, zhang2019detecting, wang2020cnn, sinitsa2024deep, marra2019gans, vahdati2023detecting, yu2019attributing, genimage_synth_det, deepfake_det_cnn_gen_facial_img, Vahdati_2024_CVPR, openset}. 

\subheader{Continual Learning For Synthetic Image Detection}
Given the distinct forensic traces left by various generator families, it is crucial for image detectors to continually update their knowledge with new generators. 
Traditional re-training is often data inefficient and fine-tuning may often lead to catastrophic forgetting~\cite{mccloskey1989catastrophic, mcclelland1995there}. Hence, a number of methods have been developed to allow the base model to adapt without losing prior knowledge~\cite{LWF, udil, der, rebuffi2017icarl}. Notably, Marra~et al.~\cite{marra2019incremental} and Kim~et al.~\cite{kim2021cored} have successfully applied such strategies to synthetic image detection, demonstrating the feasibility and effectiveness of continual learning in this domain.

%
\section{Problem Formulation}


We begin by assuming that an initial synthetic image detector $f_0$ has been created to detect images generated by a set of known generators $\gensetbl$.  We will refer to this detector as the baseline detector.  We further assume that $f_0$ was trained using a large baseline training dataset $\mathcal{B}$ consisting of both real and synthetic images 
made using generators 
in $\gensetbl$.

After the baseline detector has been trained, new synthetic image generators $g_k \notin \gensetbl$ will continue to emerge.  As previous research has shown, $f_0$ will have a difficult time detecting these new generators if the forensic traces or ``fingerprints'' left by these generators are substantially different than those left by the generators in $\gensetbl$~\cite{corvi2023detection}. 
Because of this, the synthetic image detector will need to be continually updated as new generators emerge. 

This presents an important set of 
problems. Training a new detector from scratch can be resource intensive and requires that the baseline dataset  $\mathcal{B}$ be stored indefinitely.  Instead, it is preferable to retain only a small dataset $\mathcal{M}$, known as the memory buffer, to update the detector.  
Care must be taken when updating the detector, however, because naively fine-tuning the detector can lead to catastrophic forgetting.  When this happens, the detector is able to identify images produced by the new generator, but loses its ability to reliably detect images from previous generators.

Additionally, challenges may arise when access to synthetic images generated by the new generator $g_k$ is limited.  For instance, a generator may not be available to the general public, but a small number of examples from the generator may be publicly obtainable.  
%
An example of this is OpenAI's Sora generator~\cite{sora}.
Currently, access to Sora is restricted, however OpenAI has shared a small number of sample outputs. 
Alternatively, a party engaged in a misinformation campaign may have developed a new generator for their purposes. 
In this case, access to images produced by this new generator is severely limited by the
%
number of examples that have been released into the wild.
%
%
Training an effective synthetic image detector that can continuously adapt to new generators with limited data 
is highly non-trivial.


To formalize the problem of updating the detector given these constraints, we define $\prevbuf$ as the memory buffer when the $k^{th}$ new generator is introduced. $\prevbuf$ consists of two subsets: $\realset$ which contains real images and $\prevsynthset$ which contains images made by each of the previously seen generators.
The memory buffer has a fixed size $|\mathcal{M}_\ell|=M$, that remains constant even as more generators emerge.  
%
Additionally, we define $\mathcal{D}_k$ as the set of images from the new generator $g_k$ that can be used to update the detector.  
We assume that  $|\newdata|=N$, and that $N$ is significantly smaller than the number of images in $\mathcal{B}$,
i.e. we are allowed a small number of images from the new generator.

\section{Proposed Approach}
\label{sec:proposed_approach}


While a synthetic image detector $f$ is often thought of as a single network, we can conceptually view it as the composition of an embedder $\phi$ and a classifier $h$, such that $f=h\circ\phi$. 
When adopting this view, the embeddings produced by $\phi$ capture forensic traces left by synthetic image generators, while the detector maps these embeddings to detection decisions. 
In practice, lower layers of the network can be thought of as the embedder, while the final layer or layers can be thought of as the classifier.




When a continual learning technique is used to update 
$f$, this typically involves using a special process that retrains $f$ to  detect images from both a new generator and existing generators using $\membuf$ and $\newdata$.  
This corresponds to updating $\phi$ so that it learns an embedding space that is able to jointly capture forensic traces from previous generators as well as the new generator.  
However, since forensic traces from different generators can be substantially different~\cite{corvi2023detection}, learning an embedding space that successfully does this with a limited amount of data in $\membuf$ and $\newdata$ can be challenging.


To overcome this challenge, we propose a new framework to update $f$ 
called 
\textit{\Algname} (E3). In this framework, we do not attempt to learn a single embedding space to capture the traces left by all generators.  
Instead, we form a set of expert embedders $\expset=\{\phi_0,\ldots,\phi_k\}$, where each expert embedder $\phi_\ell$ is specialized to capture traces from 
generator $g_\ell$.  
When forming an expert embedder $\phi_k$ for a new generator $g_k$, we allow it to experience catastrophic forgetting, since other experts in $\expset$ are dedicated to capturing traces from 
other generators.

To analyze an image, 
it is passed through all embedders in $\expset$ to produce a sequence of embeddings $\embseq$.  Each embedding $\mathbf{x}_\ell$ captures evidence that the image was generated by $g_\ell$.  
This set of embeddings is then analyzed using an Expert Knowledge Fusion Network (EKFN) to produce a single detection decision.  
%
As a result, E3 is able to leverage forensic evidence within 
the union of every expert's embedding space, as opposed to relying on a single embedding space to capture all forensic evidence.

We describe the process to form each expert embedder, update the memory buffer, and train the expert knowledge fusion network in the following subsections.


%


%

\begin{figure}[!t]
\centering
\includegraphics[width=0.75\linewidth]{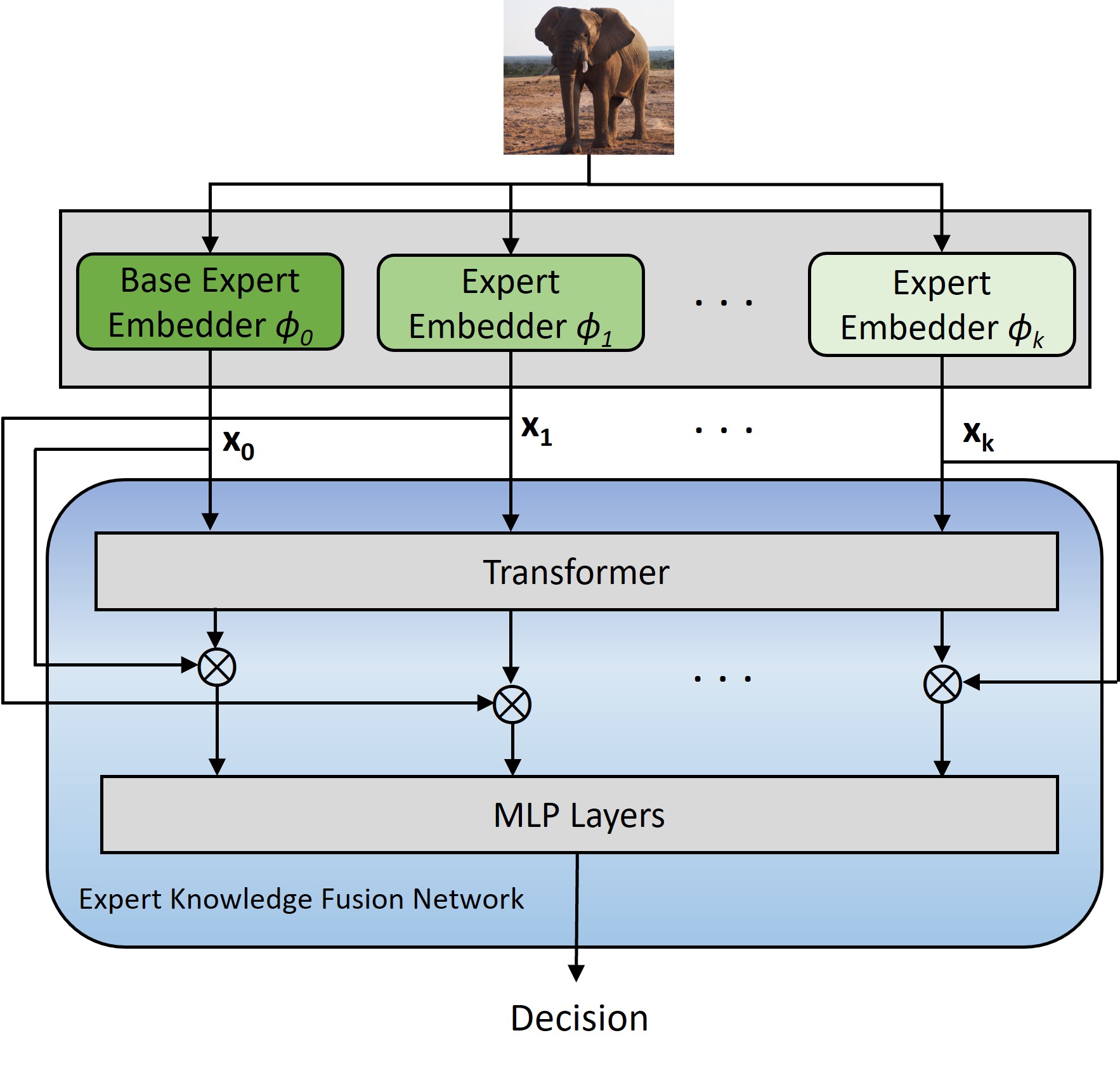}

\caption{End-to-end architecture workflow: An image is first processed by E3 to generate embeddings, which are passed through a transformer and MLP layers to produce the detection decision.}

\pulluppp\pullupp
\label{fig:overall_sys_diagram}
\end{figure}

\subsection{Creating A New Expert Embedder}
\label{subsec:new_expert_training}



\begin{figure}[!t]
    \centering
    \includegraphics[width=0.75\linewidth]{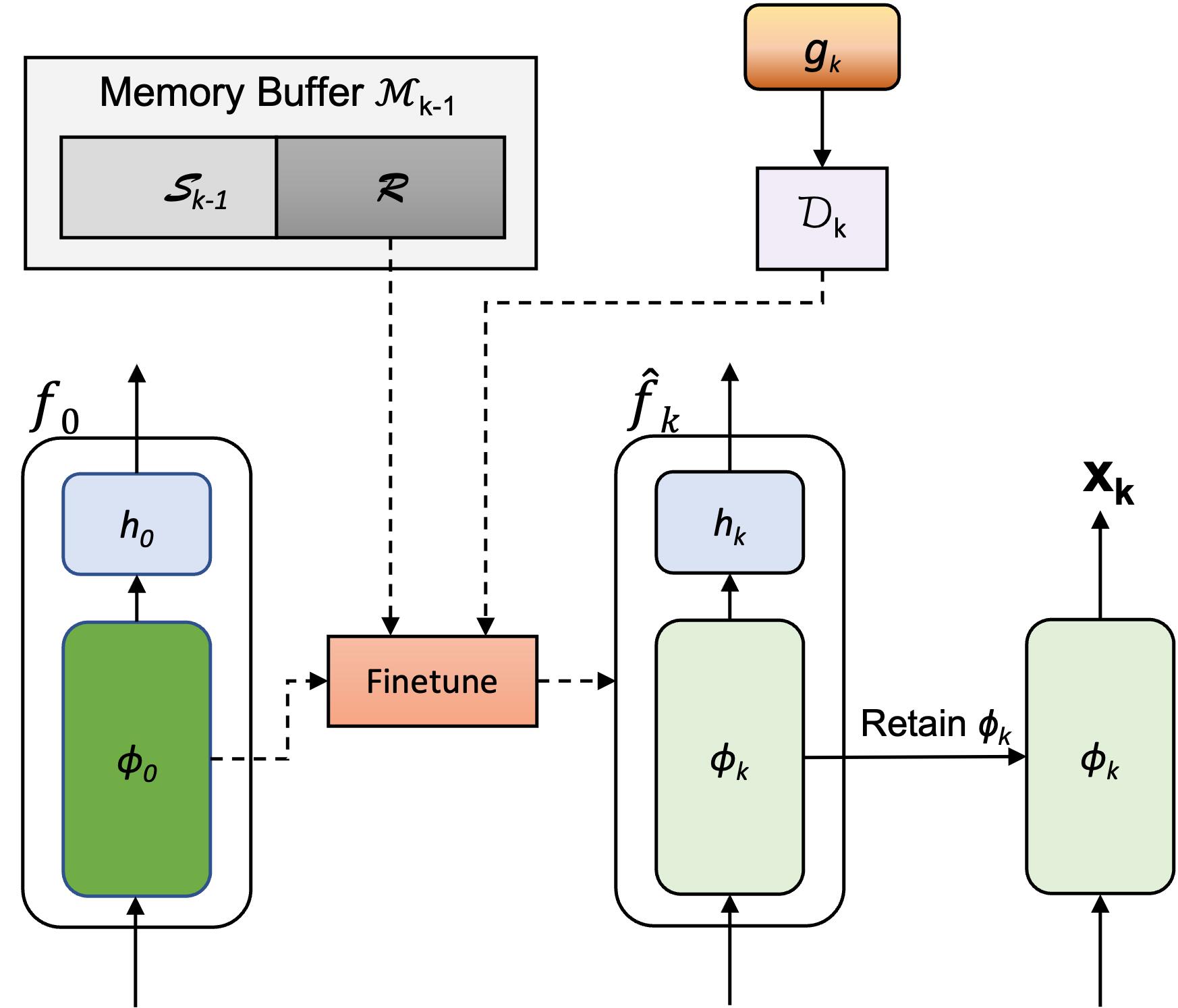}
    \caption{Creation of a new specialized expert, $\phi_k$, when a new generator emerges. The baseline detector $f_0$ is fine-tuned with data $\mathcal{D}_k$ and real images $\mathcal{R}$ from memory buffer. The new classifier $\hat{f}_k$'s embedder $\phi_k$ is then preserved.}
    \label{fig:expert_embedder_training}
    \pulluppp\pullupp
\end{figure}

Let $\mathcal{G}_{k-1}$ be the set of currently known synthetic image generators.
When a new synthetic image generator $g_k\notin\mathcal{G}_{k-1}$ emerges, we need to update our detector. 
To do this, we must first create 
a new expert embedder $\phi_k$ to capture traces left by $g_k$.  We accomplish this by adapting the baseline detector $f_0$ using transfer learning.  
An overview of this process is shown in Fig.~\ref{fig:expert_embedder_training}.


We start by forming a new expert training set $\mathcal{T}_k$ of real and synthetic images that we will use to create $\phi_k$. 
This is done by collecting a set $\mathcal{D}_k$ of $N$ synthetic images created by the new generator.
Real images are taken from the current memory buffer $\mathcal{M}_{k-1}$, which contains the subset $\mathcal{R}$ consisting of $M/2$ real images. 
As a result, the new expert training set is
$\mathcal{T}_k= \mathcal{D}_k \cup \mathcal{R}$.

Next, we use $\mathcal{T}_k$ to update $f_0$ to detect images made by $g_k$.  Specifically, we form a new detector $\hat{f}_k$ by fine tuning $f_0$  with $\mathcal{T}_k$ using the loss function 
\begin{equation}
	\resizebox{0.9\linewidth}{!}{$
		\begin{split}
			\mathcal{L}_\phi = -\!\! \sum_{I_\ell\in \mathcal{T}_k} 
			\tfrac{|\mathcal{D}_k|}{|\mathcal{T}_k|} &t_\ell log(\hat{f}_k(I_\ell)) \\[-0.75em]
			& + \tfrac{|\mathcal{R}|}{|\mathcal{T}_k|} (1 - t_\ell) log(1-\hat{f}_k(I_\ell)),
		\end{split}
	$}
\end{equation}
where $I_\ell$ is the $\ell$-th image in $\mathcal{T}_k$, and $t_\ell$ is the class label of the image $I_\ell$ where 0 means the image is real and 1 means the image is synthetic.
%
By doing this, the resulting detector $\hat{f}_k$ will be adapted to detect images made by $g_k$.  However, it will likely experience catastrophic forgetting and be poorly suited to detect generators in $\mathcal{G}_{k-1}$. 


After obtaining the expert detector $\hat{f}_k$, we decompose it into its corresponding 
embedder $\phi_k$ and classifier components. The classifier portion is discarded, while $\phi_k$ is retained. 
Finally, $\phi_k$ is added to the set of expert embedders $\Phi_k= \Phi_{k-1} \cup \{\phi_k\}$, where $\Phi_0 = \{\phi_0\}$.
%

%

\subsection{Expert Knowledge Fusion Network}
\label{subsec:ekfn_training}

After adding the new expert embedder $\phi_k$ to the set of existing \ee, we must update the Expert Knowledge Fusion Network (EKFN) $\psi$ so that it can utilize the embeddings produced by $\phi_k$.
We do this by first updating the memory buffer $\mathcal{M}_k$, then using this data to train $\psi$ to perform detection.
This process is described below.

\subheader{Updating The Memory Buffer}
When a new generator $g_k$ emerges, the current memory buffer $\prevbuf$ does not include images generated by it.  As a result, we need to add images from $\newdata$ to form the updated memory buffer $\membuf$.
However, since the memory buffer has a fixed size $M$, we must remove synthetic images made by previous generators in $\prevgenset$ to make space for these new images.

We first note that $\prevsynthset$, i.e. the subset of $\prevbuf$ that corresponds to synthetic images, contains $P_{k-1}= \lfloor M/(2k)\rfloor$ images from each generator in $\prevgenset$.  We now need to form an updated set of synthetically generated images $\synthset$ that contains an equal number of images from all generators in $\genset$.
To do this, we retain 
$P_k=(k/(k+1))P_{k-1}$ 
images corresponding to each generator in $\prevgenset$ from $\prevsynthset$ drawn uniformly at random.  These images are added to the new updated memory buffer's subset of synthetic images $\synthset$.  Finally, we add $P_k$ images drawn uniformly at random from $\newdata$ to $\synthset$ to form a complete set of synthetic images from all generators in $\genset$ for the memory buffer.  The updated memory buffer is now formed as $\membuf = \synthset \cup \realset$.

\begin{algorithm}[!t]
	\begin{algorithmic}
		\Require $\mathcal{M}_{k-1}, \mathcal{D}_{k}, \Phi_k$
		\\
		// Let $\omega_M^{(k-1)}$ be the procedure to sample images from $\mathcal{M}_{k-1}$ uniformly at random
		\\
		// Let $\omega_D^{(k)}$ be the procedure to sample images from $\mathcal{D}_{k}$ uniformly at random
		\\
		// Let $S$ be the number of training steps
		\\
		$\mathcal{M}_k \gets \omega_M^{(k-1)} (\mathcal{M}_{k-1}) \: \bigcup 
		\: \omega_D^{(k)}(\mathcal{D}_{k})$
		\For {step = $1,\cdots,S$}
		\State $\mathbf{X} \gets \{\empty\}$
		\For {$i = 1; \;\; i \leq |\mathcal{M}_k|; \;\; i \pluseq 1$}
		\State$I_i = \mathcal{M}_k\left[i\right]$; \;$\mathbf{X}_i \gets \{\empty\}$
		\For {$j = 1; \;\; j \leq |\Phi_k|; \;\; j \pluseq 1$}
		\State $\phi_{j} \gets \Phi_k\left[j\right]$
		\State $\mathbf{X}_i \gets \mathbf{X}_i \cup \{ \phi_{j}(I_i) \}$
		\EndFor
		\State $\mathbf{X} \gets \mathbf{X} \cup \{ \mathbf{X}_i \}$
		\EndFor
		\State $\mathcal{L} \gets \mathcal{L}_\psi(\mathcal{M}_k, \mathbf{X})$ \Comment{Compute loss with Eqn.~\ref{eq:l_ekfn}}
		\State $\psi \gets \text{BackPropagate}(\mathcal{L}, \psi)$ 
		\EndFor
		\\
		\Return $\psi$
	\end{algorithmic}
	\caption{\: Training \Reasoner} \label{alg:train_ekfn}
	\vspace{-0.25em}
\end{algorithm}

\subheader{Training The EKFN}
Once we have updated the memory buffer, we use $\membuf$ to train the EKFN $\psi$.  To do this, we first use the set of expert embedders $\expset$ to extract a sequence of embeddings $\embseql$ for each image $I_\ell \in \membuf$. Next, we randomly initialize $\psi$ and train it using the loss function
\skipless
\begin{equation}
	\resizebox{0.9\linewidth}{!}{$
		\label{eq:l_ekfn}
		\begin{split}
			\mathcal{L}_\psi = 
			- \!\!\sum_{I_\ell \in \mathcal{M}_k} & \psi  (\embseqlarg) \log{t_\ell} \\[-0.75em]
			& + (1 - \psi(\embseqlarg)) \log(1-t_\ell).
		\end{split}
		$}
\end{equation}

After training the EKFN, the synthetic image detector has been completely updated and the network is ready to perform detection. The EKFN's training procedure is provided in Alg.~\ref{alg:train_ekfn}.

\subheader{EKFN Architecture}
We can think of each embedder $\phi_\ell$ as a mechanism that searches for evidence of traces left by generator $g_\ell$.  The resulting embedding $\mathbf{x}_\ell$ can then be viewed as a measurement of this evidence.
%
When viewed this way, 
the purpose of the EKFN is to examine all forms of evidence in the context of one another in order to make an accurate forensic decision.
%
In light of this, we design the EKFN such that it uses a transformer's self-attention mechanism to examine the sequence of embeddings in the context of one another.

An overview of our EKFN architecture is shown in Fig.~\ref{fig:overall_sys_diagram}.  The sequence of embeddings is first passed directly to a transformer.  Each transformer output is used to weight its corresponding embedding through element-wise multiplication. 
The resulting weighted embeddings are then passed to an MLP, which makes a final detection decision.  In practice, we form the MLP using 2 layers and the transformer using 20 layers in accordance to the results of an ablation study reported on in Sec.~\ref{sec:ablations}.  We note that simpler EKFN architectures can be utilized to reduce network complexity at the cost of 
decreased performance.

\subsection{E3 Framework Summary}

E3 uses the following steps to update a synthetic image detector after the emergence of a new generator:

\begin{enumerate}
    \item Form a new training set using real images from the memory buffer and a set of images from the new generator.
    \item Utilize this training set to create an expert embedder specifically trained to capture forensic traces associated with the new generator, and incorporate it into the ensemble of expert embedders.
    \item Update the memory buffer with the new generator's data.
    \item Retrain the Expert Knowledge Fusion Network (EKFN) using the data stored in the memory buffer.
\end{enumerate}

\definecolor{Mercury}{rgb}{0.894,0.894,0.894}
\begin{table}[!t]
    \centering
    \resizebox{0.95\linewidth}{!}{%
    \begin{tblr}{
        width = \linewidth,
        colspec = {m{6mm}m{9mm}m{37mm}m{0.001mm}m{2.3mm}m{9mm}m{34.5mm}},
        column{5} = {c},
        cell{2}{1} = {r=3}{Mercury,c},
        cell{2}{2} = {Mercury},
        cell{2}{3} = {Mercury},
        cell{2}{5} = {r=11}{},
        cell{3}{2} = {Mercury},
        cell{3}{3} = {Mercury},
        cell{4}{2} = {Mercury},
        cell{4}{3} = {Mercury},
        cell{5}{1} = {r=8}{c},
        vlines,
        hline{1-2,13} = {1-3,5-7}{},
        hline{5} = {1-3}{},
    }
        & \textbf{Abbrv.} & \textbf{Generator Name} &  &  & \textbf{Abbrv.} & \textbf{Generator Name}\\
        
        \rotatebox{90}{\parbox{16mm}{\centering Baseline\\Generators}}
        & SG & StyleGAN~\cite{karras2019style} &  & 
        
        \rotatebox{90}{\parbox{45mm}{\centering New~\&~Emerging~Generators}}
        & VQD & Vec. Quant. Diff.~\cite{vqdiff} \\
        & SG2 & StyleGAN2~\cite{stylegan2}  &  &  
        & DIG & Diffusion GAN~\cite{diffusion-gan} \\
        & ProG & ProGAN~\cite{progan} &  &  
        & SG3 & StyleGAN3~\cite{stylegan3} \\
        
        \rotatebox{90}{\parbox{25mm}{\centering New~\&~Emerging\\Generators}}
        & SD1.4 & Stable Diffusion 1.4~\cite{rombach2022high}  &  &  
        & GF & GANformer~\cite{ganformer} \\
        & GLD & Glide~\cite{glide}  &  &  
        & DL2 & DALL-E 2~\cite{dalle2} \\
        & MJ & Midjourney~\cite{midjourney}  &  &  
        & LD & Latent Diffusion~\cite{rombach2022high} \\
        & DLM & DALL-E Mini~\cite{dallemini}  &  &  
        & EG3 & Eff.Geo. 3D GAN~\cite{eg3d} \\
        & TT & Taming Transformers~\cite{taming-trans}  &  &  
        & PG & Projected GAN~\cite{projectedGAN} \\
        & SD2.1 & Stable Diffusion 2.1~\cite{rombach2022high}  &  &  
        & SD1.1 & Stable Diffusion 1.1~\cite{rombach2022high} \\
        & CIPS & Cond.Indep. Pix.Synt.~\cite{cips}  &  &  
        & DDG & Denoise Diff. GAN~\cite{DDG} \\
        & BG & BigGAN~\cite{brock2018large}  &  &  
        & DDP & Denoise Diff. Prob.~\cite{ho2020denoising} 
    \end{tblr}
    }
    \caption{The full names and abbreviations for all synthetic image generators used in this paper.}
    \label{tab:gen_abv}
    \pulluppp
\end{table}


\begin{table*}[!t]
	\centering
	\resizebox{1.0\linewidth}{!}{
		\begin{tblr}{
				width = 1.0\linewidth,
				colspec = {|c|c|c|c|c|c|c|c|c|c|c|c|c|c|c|c|c|c|c|c|c|c|},
				column{1} = {l},
                row{2} = {c},
                cell{1}{1} = {r=2}{},
                cell{1}{2} = {c=19}{0.772\linewidth,c},
                cell{1}{21} = {r=2}{},
				vlines,
				hline{1,3,12-13} = {-}{},
                hline{2} = {2-20}{},
			}
			\textbf{Method} 
            & \textbf{Synthetic Image Generators} &  &  &  &  &  &  &  &  &  &  &  &  &  &  &  &  &  &  
            & \textbf{Avg. $\pm$ Std.}\\
            & SD1.4 & GLD & MJ & DLM & TT & SD2.1 & CIPS & BG & VQD & DIG & SG3 & GF & DL2 & LD & EG3 & PG & SD1.1 & DDG & DDP & \\
			Baseline  
            & 0.89  & 0.98 & 0.85 & 0.86 & 0.81 & 0.89  & 0.76 & 0.77 & 0.85 & 0.76 & 0.98 & 0.73 & 0.97 & 0.81 & 0.98 & 0.89 & 0.87 & 0.67 & 0.78 & 0.85 $\pm$ .090 \\
			Fine-Tune 
            & 0.65  & 0.83 & 0.65 & 0.67 & 0.64 & 0.69  & 0.85 & 0.75 & 0.88 & 0.88 & 0.90 & 0.92 & 0.79 & 0.68 & 0.89 & 0.76 & 0.74 & 0.87 & 0.84 & 0.78 $\pm$ .098 \\
			LWF~\cite{LWF}       
            & 0.91  & 0.98 & 0.92 & 0.82 & 0.85 & 0.92  & 0.85 & 0.88 & 0.96 & 0.94 & 0.99 & 0.95 & 0.94 & 0.72 & 0.97 & 0.92 & 0.91 & 0.95 & 0.94 & 0.91 $\pm$ .066\\
			ER~\cite{ER}        
            & 0.94  & 0.99 & 0.97 & 0.96 & 0.96 & 0.94  & 0.99 & 0.91 & 0.98 & 0.98 & 0.99 & 0.99 & 0.99 & 0.90 & 1.00 & 0.96 & 0.94 & 0.99 & 0.99 & 0.97 $\pm$ .030\\
			DER++~\cite{der}    
            & 0.94  & 0.99 & 0.96 & 0.95 & 0.95 & 0.93  & 0.98 & 0.89 & 0.97 & 0.97 & 0.99 & 0.98 & 0.99 & 0.90 & 1.00 & 0.96 & 0.93 & 0.99 & 0.98 & 0.96 $\pm$ .031\\
			UDIL~\cite{udil}      
            & 0.94  & 1.00 & 0.97 & 0.97 & 0.96 & 0.94  & 0.99 & 0.92 & 0.98 & 0.98 & 0.99 & 0.99 & 0.99 & 0.91 & 1.00 & 0.97 & 0.93 & 0.99 & 0.99 & 0.97 $\pm$ .028\\
			ICaRL~\cite{rebuffi2017icarl}     
            & 0.88  & 0.96 & 0.91 & 0.91 & 0.89 & 0.91  & 0.96 & 0.87 & 0.94 & 0.95 & 0.91 & 0.95 & 0.95 & 0.86 & 0.97 & 0.89 & 0.88 & 0.93 & 0.96 & 0.92 $\pm$ .034\\
			MT-SC~\cite{marra2019incremental}      
            & 0.91  & 0.99 & 0.93 & 0.93 & 0.92 & 0.92  & 0.98 & 0.90 & 0.96 & 0.96 & 0.97 & 0.98 & 0.97 & 0.88 & 0.99 & 0.94 & 0.91 & 0.98 & 0.98 & 0.95 $\pm$ .034\\
			MT-MC~\cite{marra2019incremental}      
            & 0.92  & 1.00 & 0.95 & 0.95 & 0.94 & 0.93  & 0.99 & 0.9.0 & 0.98 & 0.98 & 0.99 & 0.99 & 0.99 & 0.88 & 0.99 & 0.96 & 0.92 & 0.99 & 0.99 & 0.96 $\pm$ .036\\
			\textbf{Ours}      
            & \textbf{0.99} & \textbf{1.00} & \textbf{0.99} & \textbf{0.99} & \textbf{0.98} & \textbf{0.98} & \textbf{0.99} & \textbf{0.99} & \textbf{0.99} & \textbf{0.99} & \textbf{0.99} & \textbf{0.99} & \textbf{1.00} & \textbf{0.96} & \textbf{1.00} & \textbf{0.99} & \textbf{0.99} & \textbf{1.00} & \textbf{0.99} & \textbf{0.99 $\pm$ .010}
		\end{tblr}
	}
    \caption{Detection performance of each method, measured in terms of AUC, when adapting to a single new generator. The ``baseline'' detector has only seen data from the baseline detector dataset and all approaches started from the same baseline detector.}
    \label{tab:exp1_add_single_new_gen}
    \pulluppp
\end{table*}

\section{Experiments}
\label{sec:experiments}

\subsection{Experimental Setup}
\label{subsec:exp_setup}

\subheader{Baseline Detector Dataset}
We created a baseline dataset in order to train and evaluate the baseline synthetic image detectors.
To form a diverse set of real images, we gathered 72,000 images equally distributed among the CoCo dataset~\cite{coco}, the LSUN dataset~\cite{lsun}, and the CelebA dataset~\cite{celeba}. 
As for the set of synthetic images, we also gathered 72,000 GAN-generated images from the datasets used in prior work~\cite{openset, wang2020cnn}. 
The synthetic images in this set are equally distributed among three GANs: StyleGAN~\cite{karras2019style}, StyleGAN2~\cite{stylegan2}, and ProGAN~\cite{progan}. 
%
Additionally, to form the set of images used for training, validation, and testing, we split the set of real images and the set of synthetic images each into their own disjoint subsets. Specifically, we used 
132,000 images for training with 12,000 images held out for validation, and 12,000 images for testing.
%

\subheader{Emerging Generators Dataset}\label{subhead:emerging_gen_dataset}
To simulate a constantly evolving environment in which new synthetic image generators rapidly emerge, we created a dataset to train and evaluate our proposed approach in a continual learning setting.
%
To accomplish this, we first selected a diverse set of generators, composed of 19 different generation techniques. These techniques are listed in Tab.~\ref{tab:gen_abv}. All synthetic images in this dataset are assembled from publicly available datasets~\cite{openset,artifact,sinitsa2024deep}. Then, from each generator, we gathered 600 images exclusively used for training and 400 images exclusively used for testing.


\subheader{Baseline Detector Training}
We obtained a baseline detector by training MISLnet~\cite{mislnet} to distinguish between real and synthetic images in our baseline detector dataset. We chose MISLnet as the architecture for our experiments because it is lightweight and has shown repeatedly to obtain strong performance on detecting synthetic images and other image forensic tasks~\cite{Bayar_2017_ICASSP, Bayar_2016_IHMMSEC, Mayer_2020_ICASSP, FSG, MISLgan}. We trained MISLnet using a Binary Cross-Entropy Loss function via the ADAM optimizer~\cite{Adam} with a constant learning rate of $5.0 \times 10^{-5}$. This model resulted in an accuracy of 0.97 on the testing portion of our baseline detector dataset. 




\subheader{Memory Buffer}
Throughout all experiments in this paper, we fixed the size of memory buffer, $\mathcal{M}$, to 1000 images from which 500 are synthetic and 500 are real. The synthetic images are then equally distributed among all previously known generators. We note that the memory buffer is continually updated after each step to include samples of the new generator.

\subheader{Performance Metrics}
To evaluate the performance of our proposed approach and others, we chose the Area Under the ROC Curve (AUC) metric. 
Additionally, in order to obtain a more complete picture of our performance, we provided the Relative Error Reduction (RER) with respect to the second best performing method, reported as a percentage. This metric is calculated as follows:
\begin{equation}
	\text{RER} = 100 \times \frac{\text{AUC}_{N} - \text{AUC}_{R}}{1 - \text{AUC}_{R}}, 
\end{equation}
where $\text{AUC}_{R}$ is the AUC of the referencing method, and $\text{AUC}_{N}$ is the AUC of the method being compared against.


\subheader{Competing Methods}
We compared our method against multiple approaches, including two naive strategies: 
Baseline, where the baseline detector is never updated, and 
Fine-tuning, where the baseline detector is updated exclusively with data from new generators. Additionally, we benchmarked our approach against seven well-established continual learning strategies: Learning Without Forgetting (LWF)~\cite{LWF}, Experience Replay (ER)~\cite{ER}, Dark Experience Replay (DER++)~\cite{der}, Unified Domain Incremental Learning (UDIL)~\cite{udil}, Incremental Classifier and Representation Learning (ICaRL)~\cite{rebuffi2017icarl}, Multi-Task Single-Classifier (MT-SC)~\cite{marra2019incremental}, and Multi-Task Multi-Classifier (MT-MC)~\cite{marra2019incremental}. 
We benchmarked ICaRL, MT-SC, and MT-MC using our own implementations, while to benchmark the rest of the methods, we utilized the code provided by \cite{udil}.  

\begin{table*}[!t]
	\centering
	\resizebox{1.0\linewidth}{!}{
		\begin{tblr}{
				width = 1.0\linewidth,
				colspec = {|c|c|c|c|c|c|c|c|c|c|c|c|c|c|c|c|c|c|c|c|},
				column{1} = {l},
                row{2} = {c},
                cell{1}{1} = {r=2}{},
                cell{1}{2} = {c=19}{0.772\linewidth,c},
				vlines,
				hline{1,3,12-13} = {-}{},
                hline{2} = {2-20}{},
			}
			\textbf{Method} 
            & \textbf{Synthetic Image Generators} &  &  &  &  &  &  &  &  &  &  &  &  &  &  &  &  & \\
            & SD1.4 & GLD & MJ & DLM & TT & SD2.1 & CIPS & BG & VQD & DIG & SG3 & GF & DL2 & LD & EG3 & PG & SD1 & DDG & DDP \\
			Baseline
            & 0.89 & 0.91 & 0.86 & 0.83 & 0.80 & 0.79 & 0.76 & 0.73 & 0.73 & 0.71 & 0.73 & 0.71 & 0.73 & 0.72 & 0.74 & 0.74 & 0.74 & 0.72 & 0.71\\
			Fine-Tune
            & 0.65 & 0.79 & 0.73 & 0.81 & 0.83 & 0.83 & 0.80 & 0.84 & 0.82 & 0.82 & 0.85 & 0.85 & 0.73 & 0.79 & 0.81 & 0.70 & 0.76 & 0.83 & 0.74\\
			LWF~\cite{LWF}
            & 0.90 & 0.91 & 0.89 & 0.89 & 0.88 & 0.89 & 0.86 & 0.85 & 0.85 & 0.85 & 0.82 & 0.82 & 0.72 & 0.80 & 0.82 & 0.81 & 0.80 & 0.83 & 0.79\\
			ER~\cite{ER}
            & 0.94 & 0.96 & 0.96 & 0.96 & 0.95 & 0.95 & 0.95 & 0.91 & 0.93 & 0.93 & 0.94 & 0.94 & 0.93 & 0.91 & 0.93 & 0.90 & 0.90 & 0.93 & 0.93\\
			DER++~\cite{der} 
            & 0.96 & 0.96 & 0.92 & 0.91 & 0.88 & 0.89 & 0.86 & 0.84 & 0.83 & 0.82 & 0.82 & 0.82 & 0.79 & 0.81 & 0.82 & 0.82 & 0.82 & 0.82 & 0.80\\
			UDIL~\cite{udil} 
            & 0.95 & 0.97 & 0.96 & 0.96 & 0.96 & 0.95 & 0.96 & 0.93 & 0.94 & 0.94 & 0.94 & 0.94 & 0.94 & 0.93 & 0.94 & 0.92 & 0.92 & 0.93 & 0.93\\
			ICaRL~\cite{rebuffi2017icarl} 
            & 0.91 & 0.93 & 0.92 & 0.91 & 0.90 & 0.92 & 0.91 & 0.89 & 0.90 & 0.91 & 0.90 & 0.91 & 0.91 & 0.87 & 0.90 & 0.89 & 0.90 & 0.90 & 0.90\\
			MT-SC~\cite{marra2019incremental} 
            & 0.87 & 0.85 & 0.77 & 0.76 & 0.75 & 0.74 & 0.76 & 0.69 & 0.73 & 0.74 & 0.74 & 0.73 & 0.74 & 0.70 & 0.69 & 0.70 & 0.70 & 0.76 & 0.74\\
			MT-MC~\cite{marra2019incremental} 
            & 0.92 & 0.96 & 0.94 & 0.94 & 0.93 & 0.91 & 0.92 & 0.88 & 0.89 & 0.90 & 0.91 & 0.92 & 0.91 & 0.86 & 0.90 & 0.87 & 0.87 & 0.91 & 0.90\\
			\textbf{Ours} 
            & \textbf{0.99} & \textbf{0.99} & \textbf{0.99} & \textbf{0.99} & \textbf{0.99} & \textbf{0.98} & \textbf{0.98} & \textbf{0.98} & \textbf{0.98} & \textbf{0.98} & \textbf{0.97} & \textbf{0.97} & \textbf{0.97} & \textbf{0.97} & \textbf{0.97} & \textbf{0.97} & \textbf{0.97} & \textbf{0.97} & \textbf{0.97}
		\end{tblr}
	}
    \caption{Detection performance of each method, measured in terms of AUC, when sequentially adapting to a series of new generators. The ``baseline'' detector has only seen data from the baseline detector dataset and all approaches started from the same baseline detector.}
    \label{tab:exp2_add_multi_new_gen}
    \pulluppp
\end{table*}

\subsection{Adapting to One New Generator}
\label{subsec:exp1_add_single_gen}

In this experiment, we evaluated E3's ability to detect synthetic images from one new generator. 
This is important because in reality we do not know beforehand which generator we need to adapt our detector to.
As a result, we repeated this experiment for all 19 generators in our dataset.  
We then compared our method's performance to other techniques used to update the baseline detector. We note that all approaches started with the same baseline detector network. 

We present the results of these experiments in Tab.~\ref{tab:exp1_add_single_new_gen}. In this table, each column represents the performance of different algorithms adapting the baseline detector to one specific generator. The final column calculates the average AUC and standard deviation over all generators.


This experiment's results in Tab.~\ref{tab:exp1_add_single_new_gen}
show that our method achieves the best detection performance across all possible new generators with an average AUC of 0.99. Compared to the second best method, UDIL, whose an average AUC is 0.97, we obtained a significant relative error reduction of 66\%. 
This result shows that our approach is better at adapting to any one new generator than other approaches.

Additionally, we note that other competing methods' performance has standard deviations of about 3 to 10 times larger than ours. This result can be further examined by looking at certain occasions when the baseline detector experiences a significant performance drop detecting a specific unseen generator. When this happens, other approaches also tend to exhibit a similar drop in performance. For example, all competing methods have a significant drop adapting to Latent Diffusion (LD) and BigGAN (BG). Similarly, LWF experiences a performance drop on DLM, DER++, MT-SC and MT-MC on SD1.1, SD1.4 \& SD2.1, and ICaRL on SD1.1 \& SD1.4. This result is not surprising, however, because the baseline detector, which was trained on GAN, had to be adapted to detect diffusion models' images. This aligns with findings in prior work, which showed that synthetic image detectors ``cannot reliably detect images that present artifacts significantly different from those seen during training.''~\cite{corvi2023detection}

In contrast, our approach displays strong and stable performance irrespective of which generator was added. This is likely because our approach does not need to rely heavily on learning a single embedding space to capture traces of both existing and the new generators.

\subsection{Adapting to Multiple Emerging Generators}
\label{subsec:exp2_add_multi_gen}

In this experiment, we tested E3's ability to adapt to multiple sequentially emerging generators. We conducted this evaluation to mimic real world scenarios in which a detector needs to be able to adapt to detect synthetic images from each and every newly emerging generator. 
We then compared our method's performance to other dedicated continual learning techniques. 
After adapting to a new generator, we measured E3's and competing methods' performance in terms of average AUC over the current and the previous generators. We repeated this process until we exhausted all 19 emerging generators in our dataset.

We present the results of this experiment in Tab.~\ref{tab:exp2_add_multi_new_gen}.
These results show that our approach achieved the best performance irrespective of the number of new generators added.
Furthermore, we observe that after continually adapting to 19 different generators, our method obtained an average AUC of 0.97, a reduction of only 0.02 when compared to the initial performance of 0.99 average AUC. This result is a significant improvement over the second best method, UDIL, with 0.93 average AUC. 
Additionally, it can be seen from Tab.~\ref{tab:exp2_add_multi_new_gen} that as the number of new generators introduced increases, the more degraded the performance of some competing approaches becomes (\ie LWF, DER++, MT-SC). In contrast, our method retained strong performance throughout our experiment, with minimal reduction in performance.

Additionally, we note that other competing methods all experience difficulties detecting new synthetic images from certain generators. For example, LWF \& ER experienced significant performance drop when adapting to DALL-E 2 (DL2), similarly, UDIL's performance dropped adapting to BigGAN (BG), and ICaRL, MT-SC, MT-MC have substantial difficulties adapting to Latent Diffusion (LD). However, our proposed approach does not experience such issues. This suggests that by fusing the embedding spaces of all past and current expert embedders, we can gracefully adapt to any new generator while avoid having significant performance degradation as the number of generator being added increases.

\subsection{Effects of New Generator's Training Data Size}
\begin{figure}[!t]
    \centering
    \includegraphics[width=0.8\linewidth]{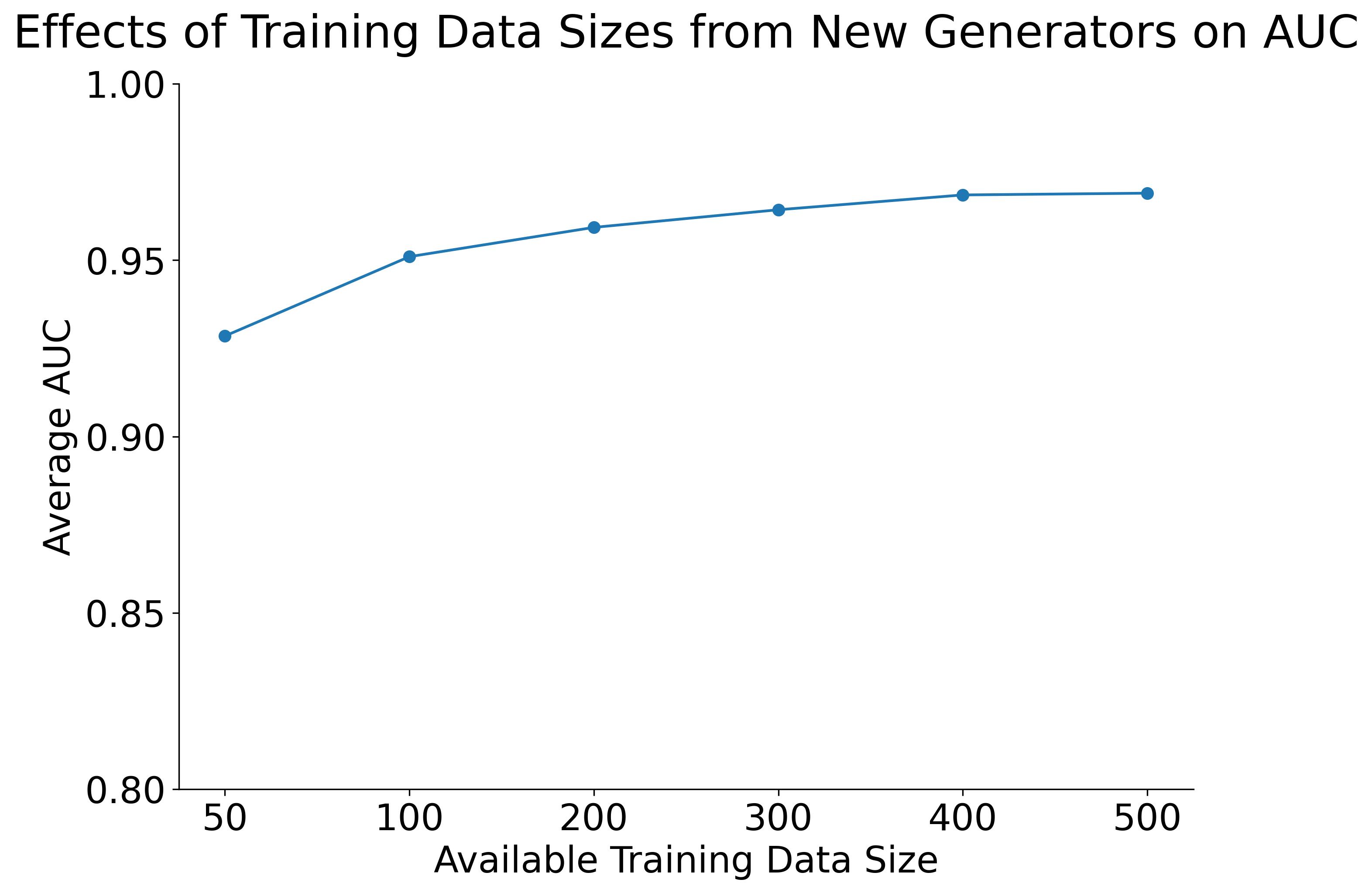}

    \caption{Our method shows slight performance decline when adapting to 19 generators with training image counts reduced from 500 to 50 per generator.}
    \label{fig:effects_new_gen_data_size}
    \pulluppp
\end{figure}


\label{subsec:exp3_new_gen_data_budget}

In this experiment, we examined the impact of different training data size from each new generator on our proposed approach. This is important because in real world scenarios, the amount of available data from a new generator is often limited, 
especially if such generator is a proprietary product.
Therefore, to conduct this experiment, we repeated the experiment in Sec~\ref{subsec:exp2_add_multi_gen}, 
with different numbers of available data ranging from 50 to 500 data points. 
We then reported the average AUC after sequentially adding all 19 generators to our dataset.

The results of this experiment are depicted in Fig.~\ref{fig:effects_new_gen_data_size}.
These results show that our method's performance experienced minimal reduction in detection performance as the number of available training images decreased. Specifically, we received a very slight drop in performance (0.97 to 0.95 average AUC) while having 5 times less training data, and a small drop in performance (0.97 to 0.94 average AUC) with 10 times less data. Notably, our performance of 0.94 average AUC is still an improvement over UDIL (0.93 AUC), whose training data size is 10 times larger than ours (see Sec.~\ref{subsec:exp2_add_multi_gen}).
This suggests that our approach not only achieves the highest performance in adapting to detect synthetic images from new generators but also requires significantly less data than other methods to attain strong performance.

\subsection{Effects of Different Detector Architectures}
\label{subsec:exp4_diff_arch}

\begin{table}[!t]
	\centering
	\resizebox{0.85\linewidth}{!}{%
    \begin{tblr}{
        width = \linewidth,
        colspec = {|c|c|c|c|c|c|},
        column{1} = {l},
        row{2} = {c},
        column{6} = {c},
        cell{1}{1} = {r=2}{},
        cell{1}{2} = {c=2}{0.204\linewidth,c},
        cell{1}{4} = {c=2}{0.204\linewidth,c},
        cell{1}{6} = {r=2}{},
        vlines,
        hline{1,3,7-8} = {-}{},
        hline{2} = {2-5}{},
    }
        \textbf{Architecture} & \textbf{Accuracy} &  & \textbf{AUC} &  & {\textbf{AUC}\\\textbf{RER}}\\
        & \textbf{Ours} & UDIL & \textbf{Ours} & UDIL & \\
        MISLnet~\cite{mislnet} & \textbf{0.922} & 0.860 & \textbf{0.970} & 0.932 & 55.9\%\\
        ResNet-50~\cite{resnet} & \textbf{0.817} & 0.790 & \textbf{0.901} & 0.890 & 10.0\%\\
        DenseNet~\cite{huang2017densely} & \textbf{0.810} & 0.800 & \textbf{0.890} & 0.869 & 16.0\%\\
        SR-Net~\cite{srnet} & \textbf{0.971} & 0.964 & \textbf{0.996} & 0.994 & 33.3\%\\
        \textbf{Average} & \textbf{0.880} & 0.853 & \textbf{0.939} & 0.921 & 22.8\%
    \end{tblr}
    }
	\caption{The performance of E3 versus UDIL using different basline architectures. AUC-RER is also provided.}
	\label{tab:exp4_archs}
    \pulluppp
\end{table}


In this experiment, we examined the effects of different network architectures for the baseline detector. 
This is to demonstrate the generality of our approach over the diverse set of detection algorithms in the wild.
To do this, we repeated the experiment in Sec.~\ref{subsec:exp2_add_multi_gen} with
three additional network architectures: ResNet-50~\cite{resnet}, SR-Net~\cite{srnet}, and DenseNet~\cite{huang2017densely}. These networks are widely used in prior work for synthetic image detection and other forensic tasks~\cite{resnet_method_2, resnet50_method_3, yousfi2020imagenet, singh2021steganalysis, kim2021cored, densenet_method3}. We then measured accuracy and AUC after adapting to all 19 generators in our dataset. We note that due to space limitation, we will only compare to UDIL, the best performing competitor in both experiments in ~\cref{subsec:exp1_add_single_gen,subsec:exp2_add_multi_gen}.

We present this experiment's results in Tab.~\ref{tab:exp4_archs}.
These results show that our proposed approach achieves the best performance irrespective of the underlying detector architecture. In particular, we notably outperform UDIL in terms of both accuracy and AUC when employing MISLnet, ResNet-50, or DenseNet as the network architectures for the baseline detector. Across all architectures, our performance is a 22.8\% relative error reduction when compared to UDIL. These results suggest that our proposed method is invariant to different architecture designs and can be applied in a general manner to most, if not all, synthetic image detectors.

\section{Discussion}
\label{sec:discussion}

Our method's strong performance can be attributed to its utilization of dedicated embedders for each new generator. These embeddings are specifically learned to capture forensic traces left by their corresponding generators. As a result, our algorithm does not rely solely on a single embedding space, but rather on the union of all embedding spaces from every expert embedder in the ensemble.

However, our approach does come at a cost of increased network size. 
Despite this, in many cases, the cost of mis-detecting AI-generated images, or false-alarming real images as synthetic, is worth the increase in network parameters. Additionally, it is important to note that after our method adapted the baseline detector (based on MISLnet) to 19 different generators, the total number of parameters in our network is 27.8M, a comparable number to a single ResNet-50 model with 23.5M parameters. Furthermore, 25.1M of these parameters are frozen because they come from the ensemble of embedders. In reality, only about 4M parameters in our network are trainable. For perspective, after adding 100 new generators, the trainable portion of our network is only 13.6M parameters.

\section{Ablation Results}
\label{sec:ablations}

\begin{table}[!t]
    \centering
    \resizebox{0.85\linewidth}{!}{%
    \begin{tblr}{
        width = \linewidth,
        colspec = {m{47mm}m{17mm}m{17mm}}, 
        column{2,3} = {c},
        vlines,
    }
        \hline
        \textbf{Setup} & \textbf{Accuracy} & \textbf{AUC} \\ 
        \hline
        \textbf{Proposed} & \textbf{0.93} & \textbf{0.97} \\ 
        \hline
        Majority Voting & 0.50 (-0.43) & 0.90 (-0.07) \\
        Knowledge Fusion w/ MLP only & 0.91 (-0.02) & 0.96 (-0.01)\\
        No Transformer Weighting & 0.88 (-0.05) & 0.96 (-0.01) \\
        5 Transformer Layers & 0.91 (-0.02) & 0.97 (-0.00)\\
        10 Transformer Layers & 0.91 (-0.02) & 0.96 (-0.01) \\      
        \hline
    \end{tblr}
    }
    \caption{Ablation study of the different design choices of the Expert Knowledge Fusion Network.}
    \label{tab:ablation}
    \pulluppp
\end{table}

We conducted an ablation study to assess the impact of different design choices of our Expert Knowledge Fusion Network (EKFN). We present these results in Tab.~\ref{tab:ablation}.

\subheader{Majority Voting}
In this setup, we fused the knowledge from each expert embedder using a majority voting strategy. This approach involves each expert embedder producing their own decisions about the input image and the final detection score is decided using a majority vote. As shown in Tab.~\ref{tab:ablation}, this approach resulted in a significant drop in both accuracy and AUC. We note that the reason why this version has an AUC of 0.90 with an accuracy of 0.50 is because it produces an uncalibrated decision score, a similar phenomenon observed in Corvi et al.~\cite{corvi2023detection}.

\subheader{Knowledge Fusion with MLP Only}
In this experiment, we removed the transformer from the EKFN and evaluated the performance of our approach. Tab.~\ref{tab:ablation} shows that this version experience a performance reduction compared to our proposed method. Hence, the transformer is important to obtain strong performance.

\subheader{No Transformer Weighting}
In this experiment setup, we discarded the weighting between the transformer output and its input. Results in Tab.~\ref{tab:ablation} show that this version achieves worse performance than our proposed method. Hence, this integration approach is important for our method.



\subheader{Number of Transformer Layers}
We tested our approach with the transformer made using only 5 or 10 layers. Tab.~\ref{tab:ablation}'s results show that this is sub-optimal to our approach in terms of AUC and accuracy.

\section{Conclusion}
\label{sec:conclusion}

This paper introduces E3, a novel approach for effectively updating synthetic image detectors to accurately detect newly emerging generators with minimal training data. By developing expert embedders tailored to capture traces from each new target generator, our method demonstrates strong adaptability. The proposed expert knowledge fusion network analyzes forensic evidence from all experts, facilitating precise detection decisions. Through extensive experimentation, E3 consistently outperforms competing continual learning approaches, even across various detector architectures and with limited data from new generators. 

\skipnormal
\subheader{Acknowledgments}
This material is based on research sponsored by DARPA and the Air Force Research Laboratory (AFRL) under agreement number HR0011-20-C-0126 and by the National Science Foundation under Award No. 2320600.

{
    \small
    \bibliographystyle{ieeenat_fullname}
    \bibliography{main.bib}
}


\end{document}